\documentclass[conference]{IEEEtran}
\IEEEoverridecommandlockouts
\usepackage{cite}
\usepackage{amsmath,amssymb,amsfonts}
\usepackage{algorithmic}
\usepackage{graphicx}
\usepackage{textcomp}
\usepackage{xcolor}

\usepackage{subfigure}
\usepackage{makecell}
\usepackage{multirow}
\usepackage[hidelinks]{hyperref}
\usepackage{color}
\def\BibTeX{{\rm B\kern-.05em{\sc i\kern-.025em b}\kern-.08em
    T\kern-.1667em\lower.7ex\hbox{E}\kern-.125emX}}
\begin{document}

\title{Knowledge Distillation from 3D to Bird's-Eye-View for LiDAR Semantic Segmentation
\thanks{This work is supported by Shanghai Municipal Science and Technology Major Project (No. 2018SHZDZX01), ZJLab, Shanghai Center for Brain Science and Brain-Inspired Technology.}
}

\author{\IEEEauthorblockN{Feng Jiang}
\IEEEauthorblockA{\textit{ISTBI} \\
\textit{Fudan University}\\
Shanghai, China \\
jiangf21@m.fudan.edu.cn}

\\

\IEEEauthorblockN{Haiqiang Zhang}
\IEEEauthorblockA{\textit{Mogo Auto}\\
Beijing, China \\
zhanghaiqiang@zhidaoauto.com}

\and 

\IEEEauthorblockN{Heng Gao}
\IEEEauthorblockA{\textit{ISTBI} \\
\textit{Fudan University}\\
Shanghai, China \\
hgao22@m.fudan.edu.cn}
\\
\IEEEauthorblockN{Ru Wan}
\IEEEauthorblockA{\textit{Mogo Auto}\\
Beijing, China \\
wanru@zhidaoauto.com}

\and

\IEEEauthorblockN{Shoumeng Qiu}
\IEEEauthorblockA{\textit{School of Computer Science} \\
\textit{Fudan University}\\
Shanghai, China \\
skyshoumeng@163.com}

\\

\IEEEauthorblockN{Jian Pu$^\ast$}
\IEEEauthorblockA{\textit{ISTBI, Fudan University}\\
Shanghai, China \\
jianpu@fudan.edu.cn}
}

\maketitle

\begin{abstract}
LiDAR point cloud segmentation is one of the most fundamental tasks for autonomous driving scene understanding. However, it is difficult for existing models to achieve both high inference speed and accuracy simultaneously.
For example, voxel-based methods perform well in accuracy, while Bird's-Eye-View (BEV)-based methods can achieve real-time inference. To overcome this issue, we develop an effective 3D-to-BEV knowledge distillation method that transfers rich knowledge from 3D voxel-based models to BEV-based models. Our framework mainly consists of two modules: the voxel-to-pillar distillation module and the label-weight distillation module. Voxel-to-pillar distillation distills sparse 3D features to BEV features for middle layers to make the BEV-based model aware of more structural and geometric information. Label-weight distillation helps the model pay more attention to regions with more height information. Finally, we conduct experiments on the SemanticKITTI dataset and Paris-Lille-3D. The results on SemanticKITTI show more than 5\% improvement on the test set, especially for classes such as motorcycle and person, with more than 15\% improvement. The code can be accessed at \href{https://github.com/fengjiang5/Knowledge-Distillation-from-Cylinder3D-to-PolarNet}{https://github.com/fengjiang5/Knowledge-Distillation-from-Cylinder3D-to-PolarNet}.
\end{abstract}


\section{Introduction}
\label{sec:intro}
The 3D point clouds can effectively capture the real-world scene while preserving geometric and structural information to the greatest extent. Point cloud semantic segmentation plays a very important role in the perception of the surrounding environment, especially in autonomous driving, which aims to predict a label for each point in the current scan~\cite{behley2019iccv}. Therefore, balancing speed and accuracy is especially important.

Currently, deep learning-based LiDAR segmentation methods can be generally categorized as point-based~\cite{qi2017pointnet, thomas2019kpconv, hu2020randla}, projection-based~\cite{zhang2020polarnet, milioto2019rangenet++, wu2019squeezesegv2, cortinhal2020salsanext} and voxel-based~\cite{zhu2021cylindrical, hou2022point}. For instance, PointNet~\cite{qi2017pointnet} is a point-based method that only uses a stack of MLPs to learn the representations of raw point clouds directly. 
Projection-based methods can be further divided into two categories: range-based methods that use spherical projection~\cite{milioto2019rangenet++, cortinhal2020salsanext} and Bird's-Eye-View (BEV)-based methods~\cite{zhang2020polarnet}. After point cloud projection, these methods can take full advantage of traditional 2D convolutional neural networks. The inference speed of these methods is very fast and can meet the requirements of real-time~\cite{alonso20203d}, but the precision of these models is not satisfactory. In contrast, voxel-based methods~\cite{zhu2021cylindrical, hou2022point} can achieve high accuracy but are difficult to apply in practice due to the use of time-consuming and computationally expensive sparse operators~\cite{graham20183d}. 
Geoffrey Hinton proposed knowledge distillation~\cite{hinton2015distilling}, which is usually used for model compression. It trains the simple student model by using the outputs of the teacher model as a supervisory signal. Currently, many methods used for point cloud segmentation apply knowledge distillation for model compression~\cite{hou2022point} or better feature extraction~\cite{yan20222dpass}. However, few methods focus on knowledge distillation between two different point cloud segmentation methods.

Therefore, the trade-off between high accuracy and high inference speed is a pivotal problem for practical applications, such as autonomous driving. To address this issue, we propose a novel framework to improve the accuracy of BEV-based models while maintaining their real-time inference speed.
We develop voxel-to-pillar distillation and label-weight distillation modules for knowledge distillation. 
Specifically, voxel-to-pillar distillation is an attention-based method for middle feature distillation that utilizes cross-attention mechanisms and MSE loss to help BEV-based models learn more structural and spatial geometric information from voxel-based models. Label-weight distillation is used for the last layer before classification, which selects regions with relatively large values in the height count map for distillation since these regions are the main reason for the performance gap. 

We evaluate our model on the SemanticKITTI dataset~\cite{behley2019iccv} and Paris-Lille-3D~\cite{roynard2017parislille3d}, and it outperforms the original PolarNet~\cite{zhang2020polarnet} by 5\% mIoU on the test set of SemanticKITTI. We also conduct ablation studies to demonstrate the effect of the distillation modules. The results show that our framework can largely improve the segmentation performance of PolarNet, especially the performance for classes that have more height information, such as motorcycles and persons whose heights along the z-axis are compressed under the Bird's-Eye-View. 

The main contributions of our work are threefold:
\begin{itemize}
  \item We propose an effective knowledge distillation framework that transfers knowledge from 3D voxel-based to BEV-based models for LiDAR point cloud segmentation.
  \item We develop a novel voxel-to-pillar distillation module that helps BEV-based models learn more deep structural and spatial geometric information. Moreover, to reduce the height information loss, we design a label-weight distillation module focusing more on key regions with rich height information.
  \item We conduct experiments on the SemanticKITTI dataset. The results show that our framework can outperform the original by 5\% mIoU on the test set, especially in classes such as motorcycle and person, with more than 15\% improvement. We also conduct experiments on Paris-Lille-3D, and the results demonstrate our method's effectiveness and generalization power.
\end{itemize}

\section{Related Work}
\subsection{LiDAR Segmentation}

Currently, LiDAR point cloud segmentation methods~\cite{zhang2020polarnet, milioto2019rangenet++, wu2019squeezesegv2, zhu2021cylindrical, wu2018squeezeseg, xu2020squeezesegv3}, which are crucial for autonomous driving scene understanding, can be mainly separated into three categories: point-based, projection-based and voxel-based. Point-based methods, such as PointNet~\cite{qi2017pointnet} and KPConv~\cite{thomas2019kpconv}, consume raw point clouds as input without any voxelization or other intermediate representations, which is straightforward. However, these methods are computationally intensive due to a large number of laser points. 

To overcome this problem, methods based on range view (RV) and bird’s-eye-view (BEV) first project 3D points into 2D space and then feed the projected maps into a 2D convolution network for feature extraction. For instance, PolarNet~\cite{zhang2020polarnet} quantizes the points into polar BEV grids and uses ring CNNs for downstream segmentation tasks, which improves accuracy while maintaining real-time inference speed. However, projection-based methods will inevitably lose considerable geometric information due to dimension compression. Cylinder3D~\cite{zhu2021cylindrical}, a classic voxel-based method, proposes to use a cylindrical partition and an asymmetrical 3D sparse module to retain 3D representation and tackle the issues induced by the point cloud's sparsity and unevenness of distribution at the same time. Although voxel-based methods have the ability to achieve high accuracy, they demand a huge amount of memory usage and computing resources. Therefore, they are difficult to use in real-world applications such as real vehicle deployment. To reach high precision and decrease inference time, we propose a novel knowledge distillation framework to help BEV-based models learn more geometric and structural information.


\subsection{Knowledge Distillation}

Knowledge distillation, first proposed by Geoffrey Hinton in \cite{hinton2015distilling}, is a generic model compression architecture that transfers the outstanding learning representation capability of cumbersome and large models to small but easily deployable models. In addition, it can also be used for auxiliary supervision when training the model~\cite{yan20222dpass, wu2022attention}.

\noindent Wu \textit{et al.}~\cite{wu2022attention} proposed ADD, an attention-based depth knowledge distillation framework with a 3D-aware positional encoding that uses an extra lidar branch to capture depth information. They used cross-attention to improve training from the lidar branch to the main branch, and the results showed that this is better than the simple use of MSE loss.
Hou~\cite{hou2022point} claimed that their Point-to-Voxel Knowledge Distillation (PVKD) is the first work that applied knowledge distillation for LiDAR semantic segmentation. In their work, they proposed a supervoxel partition method to divide the point clouds into several supervoxels and designed a difficulty-aware sampling strategy to more frequently sample supervoxels containing less-frequent classes and faraway objects.
Yan~\cite{yan20222dpass} proposed a 2D priors assisted semantic segmentation method to boost the representation learning on point clouds. They have achieved good results but are still limited by speed because of sparse operators. 

The previous methods rarely involved knowledge distillation from voxel-based methods to BEV-based methods. Our proposed framework can solve this problem well, especially for classes where much information is lost during projection, such as motorcycles and persons.

\section{Method}
\label{sec:method}

\begin{figure}[t]
\centering
\includegraphics[width=1\linewidth]{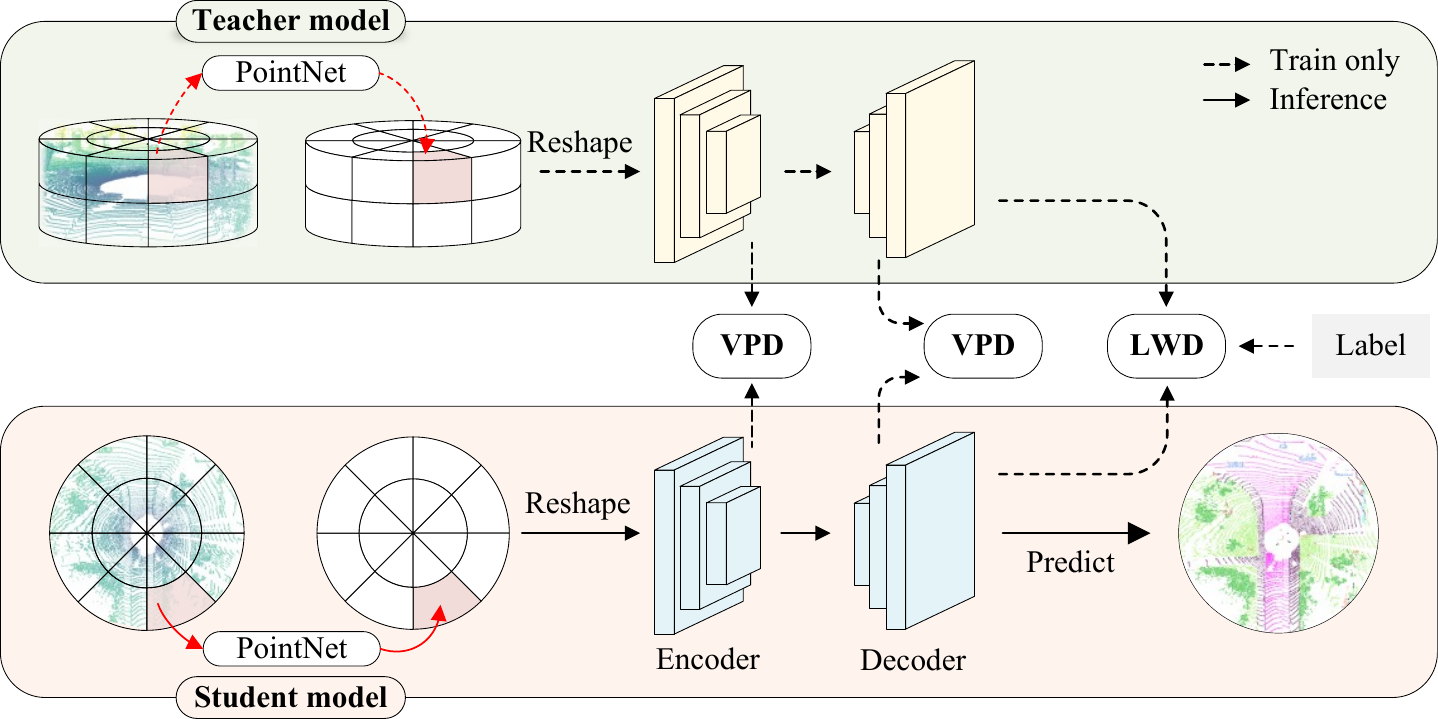}

\caption{The overview of our framework consists of three main parts: the teacher model, the student model and the knowledge distillation model. During training, we use pretrained weights for the teacher model and update the parameters of the student model and knowledge distillation model. Voxel-to-pillar distillation (VPD) is used for general middle layers, and label-weight distillation (LWD) is specifically used for the layer before classification.}
\label{fig:framework}
\end{figure}

Given a certain scan of point clouds $\mathbf{P} \in \mathbb{R}^{N \times 4}$, the objective of semantic segmentation is to predict a label for each point, where $N$ is the number of points. Each point contains ($x,y,z,i$), where $(x,y,z)$ are the Cartesian coordinates relative to the LiDAR scanner and $i$ is the reflection intensity.

We propose a novel framework that can transfer knowledge from accurate voxel-based methods to efficient BEV-based methods through knowledge distillation. Specifically, we propose two general modules for knowledge distillation from 3D sparse features to BEV features to reduce the gap between these two methods. First, the point clouds independently pass through the voxel-based methods and the BEV-based methods. Then voxel-to-pillar distillation and label-weight distillation are used for certain layers to help the BEV-based methods learn more geometric and structural information. The details of the proposed framework are as follows.

\subsection{Framework Overview}
The architecture of our framework is shown in Fig.~\ref{fig:framework}. Our framework has three main parts, similar to other knowledge distillation methods~\cite{hou2022point, yan20222dpass, wu2022attention}. The first is the teacher model, and here, we choose voxel-based methods as our teacher model, which have high accuracy but suffer from a computational burden. Voxel-based methods usually encode each voxel and have more geometric and structural features. We choose BEV-based methods as the student model because they are usually efficient for practical application. The last part is the knowledge distillation model, which is the main part of our framework. Benefiting from our well-designed voxel-to-pillar distillation module and label-weight distillation module, the student model can learn more valuable information during training and improve the inference performance without extra computation. In addition, we also used logit distillation similar to others~\cite{hou2022point, hinton2015distilling}.

\begin{figure}[t]
\centering
\includegraphics[width=1\linewidth]{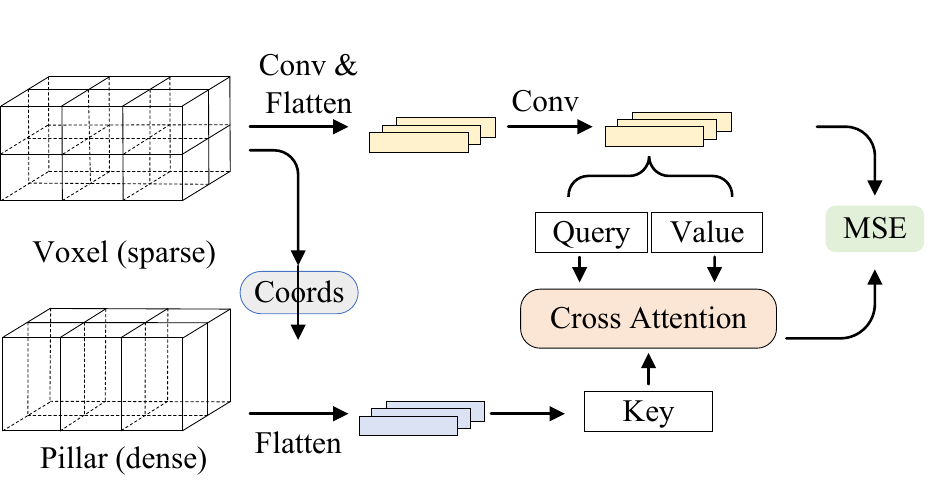}
\caption{Illustration of the voxel-to-pillar distillation (VPD) module. We flatten the 3D sparse features of the teacher model and use MLPs to transfer the domain to match the corresponding features of the student model. Then, we use a cross-attention module between the 3D sparse features and the BEV features to reinforce the learning capabilities of the student model.}
\label{fig:PVD}
\end{figure}

\subsection{Voxel-to-Pillar Distillation}
 Voxel-to-pillar distillation is one of our framework's main modules of knowledge distillation, as illustrated in Fig.~\ref{fig:PVD}. Generally, voxel-based methods use sparse operators~\cite{graham20183d} because the point clouds are sparse and unevenly distributed. To maintain the sparsity of the point clouds and reduce the computational cost, most sparse operators make the features of intermediate layers still sparse, making voxel-based methods slow and hard to align directly with BEV-based methods. BEV-based methods take pillars as pseudo images, so they can directly use efficient CNNs designed for image processing, but the information is lost at the data preprocessing stage. Our proposed module helps BEV-based methods overcome this problem to some extent.

The feature of voxel-based methods is denoted as $F_{V}^{i}$, and the corresponding BEV-based feature is denoted as $F_{B}^{i}$, where $i$ means the $i$-th layer and $F_{B}^{i} \in \mathbb{R}^{N \times C_V},F_{V}^{i} \in \mathbb{R}^{N \times C_B}$. For simplicity, here, we assume that $F_{V}^{i}$ and $F_{B}^{i}$ have the same size in the $x$-$y$ plane.
First, we use sparse convolution to compress the features of voxel-based methods $F_{V}^{i}$ in the z-axis and denote it as $F_{VC}^{i}$, where $F_{VC}^{i}=f(F_{V}^{i})$ and $f:\mathbb{R}^{N \times C_V} \rightarrow \mathbb{R}^{N \times C_B}$. Since $F_{VC}^{i}$ is sparse, we need to obtain the coordinates of non-empty voxels, which are used to match the features of BEV-based methods exactly. 
Second, we flatten features from voxel-based methods and BEV-based methods, which are denoted as $f_{V}^{i}$ and $f_{B}^{i}$. $f_{V}^{i}$ and $f_{B}^{i}$ are from different methods, so they are in different domain spaces and usually have different distributions. 
To solve this problem, we use the domain transfer method learned from BYOL~\cite{grill2020bootstrap}, specifically two MLPs with normalization. 
Specifically,
\begin{equation}
    \begin{aligned}
        f_{V}^{i} =& \text{MLP}(F(C(F_{V}^{i})))\\
        f_{B}^{i} =& F(F_{B}^{i}),
        \label{equ:cross}
    \end{aligned}
\end{equation}
where $\text{MLP}$ means domain transfer, $F$ means flattening of the features and $C$ means compression of the shape of the features.
Then, we produce the Key from $f_{B}^{i}$ and obtain the Query and Value from $f_{V}^{i}$ by MLPs. We integrate these in a cross-attention module by 
\begin{equation}
Attention(Q,K,V)=\text{softmax}(\frac{QK^T}{\sqrt{d_k}})V
\end{equation}
such as \cite{wu2022attention} to generate student features $f_{B'}^{i}$ and use MSE loss between $f_{B'}^{i}$ and $f_{V}^{i}$ to make the student methods learn the lost information during data preprocessing. We normalize the features $f_{B'}^{i}$ and $f_{V}^{i}$ learned from BYOL~\cite{grill2020bootstrap}, which means that we focus more on directional differences than numerical differences. Therefore, the point-to-voxel distillation loss is given as follows:
\begin{equation}
\mathcal{L}_{VPD}^{}\left(f_{B'}^{}, f_{V}^{}\right)=\frac{1}{|\mathcal{I}|}\sum_{i\in \mathcal{I}}\frac{1}{N_i}\left\|\frac{f_{B'}^{i}}{\left\|f_{B'}^{i}\right\|_2}-\frac{f_{V}^{i}}{\left\|f_{V}^{i}\right\|_2}\right\|_2^2
\end{equation}
where $N_i$ is the number of nonempty voxels and $\mathcal{I}$ is the subset of layers used for distillation.

\subsection{Label-Weight Distillation}
Label-weight distillation is used for the last layer before classification. One of the main reasons for the performance degradation of the BEV-based methods is that they lose considerable height information, which plays an important role, especially for hard labels and hard scenes. Fig.~\ref{visual} in Sec.~\ref{sec:visualization} demonstrates this point of view. 
When we project the points onto a $x$-$y$ plane, there may only be $4\sim6$ pillars belonging to the class person, and the class road will mislead the model because the proportion of points on the road surface is large, although these pillars belong to the class person. 
We cannot use voxel-to-pillar distillation directly because we need global information in the middle layers, so we propose label-weight distillation, which focuses on the key regions. The last layer is more sensitive to height loss than the middle layers, so we use height embedding to record the height information better. 

\begin{figure}[t]
\centering
\includegraphics[width=1\linewidth]{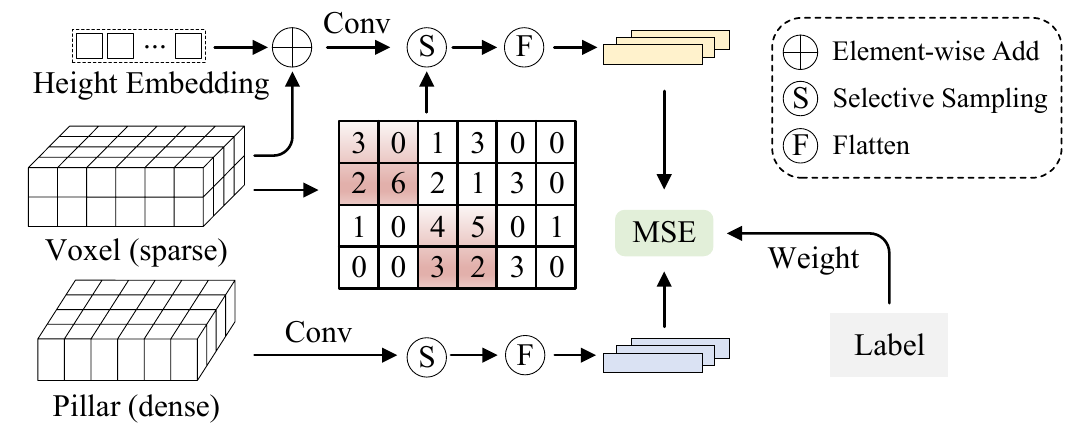}
\caption{Illustration of the label-weight distillation (LWD) module. We combine height embeddings with the original teacher features and choose features of the corresponding location using proposed selective sampling. Finally, we apply MSE loss with label-generated weights between selected features from the teacher and the student models.}
\label{fig:LWD}
\end{figure}

Fig.~\ref{fig:LWD} shows the details of label-weight distillation. Because we want the model to focus on some key regions and the last layer always has a large size, selecting some regions for distillation is necessary. We divide the whole scene into $K$ regions. According to the above analysis, we design a method to select regions with more non-empty voxels along the z-axis. Specifically, we count the height map $H$ for sparse voxel features, and the value is the number of nonempty voxels in the current position. Then, calculate weights for each region according to their ground truth label. The weight and the probability of selecting the $i$-th region are defined as $W_i=H_i/\text{sum}(H)$, $P_i=W_i/\text{sum}(W)$, where $i\in \{1\cdots K\}$.
We note that we do not select the region with no height information. Our proposed method is more suitable for distillation from voxel-based methods to BEV-based methods, especially for the height loss situation.

Specifically, we first add the height embedding $F_h$ with features from voxel-based methods $f_{V}$, which enables the model to encode deeper high-level information. We use two sparse convolution layers on $f_{V}$ after height encoding to compress the 3D features and transfer the domain to match the features of the BEV-based methods simultaneously. Then, we select $M$ regions for features from BEV-based methods $f_{B}$ and features $f_{V'}$ after compression. Finally, we flatten the features and use the MSE loss between $f_{B}$ and $f_{V^{'}}$ with weights from labels. The label-weight distillation loss is given as follows:
\begin{equation}
\mathcal{L}_{LWD}\left(f_{B}, f_{V^{'}}\right)=\frac{1}{\sum_{j=1}^M{N_j}}\left\|\frac{f_{B}^{j}}{\left\|f_{B}^{j}\right\|_2}-\frac{f_{V^{'}}^{j}}{\left\|f_{V^{'}}^{j}\right\|_2}\right\|_2^2
\end{equation}
where $j$ represents the $j$-th selected region and $N_j$ represents the number of nonempty voxels in the $j$-th region.

\subsection{Loss Function}
Our loss function has two parts: the original segmentation loss of BEV-based methods $\mathcal{L}_{\text {wce }}$ and $\mathcal{L}_{\text {lovasz }}$, the loss of distillation $\mathcal{L}_\text{{VPD}}$, the loss of label-weight distillation $\mathcal{L}_{\text{LWD}}$ and the loss of logit distillation $\mathcal{L}_{\text{Logit}}$.

$\mathcal{L}_{\text {wce }}$ means weighted cross-entropy loss. We determine the weight according to the reciprocal proportion of different categories of points in the whole training set. We assume that there are $C$ classes, $x$ is the input, $y$ is the target, $w=[w_0,,w_1,\cdots,w_{C-1}]$ is the class weight, and $N$ is the point number of current batch. 
\begin{equation}
\mathcal{L}_{\text {wce }} =- \sum_{n=1}^N \frac{1}{\sum_{n=1}^N w_{y_n} }w_{y_n} \log \frac{\exp \left(x_{n, y_n}\right)}{\sum_{c=1}^C \exp \left(x_{n, c}\right)}.
\end{equation}
$\mathcal{L}_{\text {lovasz }}$ is proposed by ~\cite{berman2018lovasz}, which is directly used on the intersection-over-union (IoU) score. Given the labels $y$, the predicts $x$, the IoU of class $c$ is defined as 
\begin{equation}
J_c\left(y, x\right)=\frac{\left|\left\{y=c\right\} \cap\{x=c\}\right|}{\left|\left\{y=c\right\} \cup\{x=c\}\right|}.
\end{equation}
\begin{equation}
\mathcal{L}_{\text {lovasz }} = 1-J_c\left(y, x\right).
\end{equation}
$\mathcal{L}_{\text {Logit}}$ is used in the probability of prediction like \cite{hinton2015distilling}.
Thus, we can obtain the final loss by
\begin{equation}
\begin{aligned}
\mathcal{L}=&\mathcal{L}_{\text {wce }}+\mathcal{L}_{\text {lovasz }}+\beta_1 \mathcal{L}_\text{{VPD}}\left(f_{B'}^{i}, f_{V'}^{i}\right)\\
&+\beta_2 \mathcal{L}_{\text{LWD}}\left(f_{B}, f_{V}\right) + \beta_3 \mathcal{L}_{\text{Logit}},
\end{aligned}
\end{equation}
where $\beta_1,\beta_2, \beta_3$ are the balance coefficients.

\section{Experiment}

\subsection{Datasets and Metrics}
SemanticKITTI~\cite{behley2019iccv} is a large dataset for LiDAR point cloud semantic segmentation, which provides annotations for all points. It has 22 point cloud sequences, each collected from a different scene. As recommended, we use 00 to 10 for training, 08 for validation during training, and 11 to 21 for testing.

We use the evaluation metric of the mean intersection-over-union (mIoU) over all classes. The IoU is defined as $m I o U=\frac{1}{n} \sum_{c=1}^n \frac{T P_c}{T P_c+F P_c+F N_c}$, where $TP_{c}$ denotes the number of true positive points for class $c$,
$FP_c$ denotes the number of false positives, and $FN_c$ is the number
of false negatives. The mIoU is the average of all classes.

\subsection{Implementation Details}

We use Cylinder3D\cite{zhu2021cylindrical} as an example of voxel-based methods and PolarNet\cite{zhang2020polarnet} for BEV-based methods. Cylinder3D uses cylindrical partition and sparse convolution network, while PolarNet first projects the point clouds to BEVs and uses ring convolution designed for BEV-based methods. They use a similar pipeline for data preprocessing, point feature extraction with PointNet\cite{qi2017pointnet}, and encoder and decoder modules.

\begin{table*}[htbp]
  \centering
  \scriptsize
  \caption{Quantitative comparison of PolarNet\cite{zhang2020polarnet} trained with our framework. The results are reported in terms of the mIoU on the SemanticKITTI test set\cite{behley2019iccv}, and $*$ indicates that the result is from the original paper. We obtain \textbf{56.1} mIoU with the officially released code. The first two groups in the table are point-based and projection-based (including BEV-based and range-based) models. The last group includes our teacher model Cylinder3D\cite{zhu2021cylindrical}, the original student model PolarNet~\cite{zhang2020polarnet} and the student model trained by our framework.}
    \setlength{\tabcolsep}{1.2mm}{
    \begin{tabular}{c|c|ccccccccccccccccccc|cc}
    \hline
    & & & & & & & & & & & & & & & & & & & &\\
    & & & & & & & & & & & & & & & & & & & &\\
    \multicolumn{1}{l|}{\makecell[c]{Categories}} & \multicolumn{1}{c|}{Methods} & {\rotatebox{90}{car}} & {\rotatebox{90}{bicycle\hspace{-1cm}}} & {\rotatebox{90}{motorcycle\hspace{-1cm}}} & {\rotatebox{90}{truck\hspace{-1cm}}} & {\rotatebox{90}{other-vehicle\hspace{-1cm}}} & {\rotatebox{90}{person\hspace{-1cm}}} & {\rotatebox{90}{bicyclist\hspace{-1cm}}} & {\rotatebox{90}{motorcyclist\hspace{-1cm}}} & {\rotatebox{90}{road}} & \multicolumn{1}{l}{\rotatebox{90}{parking}} & {\rotatebox{90}{sidewalk}} & {\rotatebox{90}{other-ground\hspace{-1cm}}} & {\rotatebox{90}{building\hspace{-1cm}}} & {\rotatebox{90}{fence\hspace{-1cm}}} & {\rotatebox{90}{vegetation\hspace{-1cm}}} & {\rotatebox{90}{trunk\hspace{-1cm}}} & {\rotatebox{90}{terrain\hspace{-1cm}}} & {\rotatebox{90}{pole\hspace{-1cm}}} & {\rotatebox{90}{traffic-sign\hspace{-1cm}}} & \multicolumn{1}{l}  {mIoU} & \multicolumn{1}{l}  {FPS}\\    \hline

    \multirow{4}{*}{Point} &\multicolumn{1}{c|}{PointNet~\cite{qi2017pointnet}} &   46.3 & 1.3 & 0.3 & 0.1 & 0.8 & 0.2 & 0.2 & 0.0 & 61.6 & 15.8 & 35.7 & 1.4 & 41.4 & 12.9 & 31.0 & 4.6 & 17.6 & 2.4 & 3.7 & 14.6 & 2\\
    
     &\multicolumn{1}{c|}{LatticeNet~\cite{rosu2019latticenet}} &   92.9 & 16.6 & 22.2 & 26.6 & 21.4 & 35.6 & 43.0 & 46.0 & 90.0 & 59.4 & 74.1 & 22.0 & 88.2 & 58.8 & 81.7 & 63.6 & 63.1 & 51.9 & 48.4 & 52.9 & 7\\
    

     &\multicolumn{1}{c|}{RandLa-Net~\cite{hu2020randla}} &  94.2 & 26.0 & 25.8 & 40.1 & 38.9 & 49.2 & 48.2 & 7.2 & 90.7 & 60.3 & 73.7 & 20.4 & 86.9 & 56.3 & 81.4 & 61.3 & 66.8 & 49.2 & 47.7 & 53.9& 22\\
    
     &\multicolumn{1}{c|}{KPConv~\cite{thomas2019kpconv}} &    96.0 & 30.2 & 42.5 & 33.4 & 44.3 & 61.5 & 61.6 & 11.8 & 88.8 & 61.3 & 72.7 & 31.6 & 90.5 & 64.2 & 84.8 & 69.2 & 69.1 & 56.4 & 47.4 & 58.8 & 38\\
    
    \hline
    
    \multirow{5}{*}{\makecell[c]{Projection}} &\multicolumn{1}{l|}{SqueezeSegV2~\cite{wu2019squeezesegv2}} & 82.7  & 21.0    & 22.6  & 14.5  & 15.9  & 20.2  & 24.3  & 2.9   & 88.5  & 42.4  & 65.5  & 18.7  & 73.8  & 41.0 & 68.5  & 36.9  & 58.9  & 12.9  & 41.0    & 39.6 & 83\\
    
    &\multicolumn{1}{c|}{RangeNet53~\cite{milioto2019rangenet++}} 
     & 91.4  & 25.7  & 34.4  & 25.7  & 23.0    & 38.3  & 38.8  & 4.8   & 91.8  & 65.0    & 75.2  & 27.8  & 87.4  & 58.6  & 80.5  & 55.1  & 64.6  & 47.9  & 55.9  & 52.2 & 12\\

    &\multicolumn{1}{c|}{SqueezeSegV3~\cite{xu2020squeezesegv3}} &  92.5 & 38.7 & 36.5 & 29.6 & 33.0 & 45.6 & 46.2 & 20.1 & 91.7 & 63.4 & 74.8 & 26.4 & 89.0 & 59.4 & 82.0 & 58.7 & 65.4 & 49.6 & 58.9 & 55.9 & 6\\

    &\multicolumn{1}{c|}{3D-MiniNet~\cite{alonso20203d} } &  90.5 & 42.3 & 42.1 & 28.5 & 29.4 & 47.8 & 44.1 & 14.5 & 91.6 & 64.2 & 74.5 & 25.4 & 89.4 & 60.8 & 82.8 & 60.8 & 66.7 & 48.0 & 56.6   & 55.8 & 28\\

    &\multicolumn{1}{c|}{SalsaNext~\cite{cortinhal2020salsanext} } &  91.9 & 48.3 & 38.6 & 38.9 & 31.9 & 60.2 & 59.0 & 19.4 & 91.7 & 63.7 & 75.8 & 29.1 & 90.2 & 64.2 & 81.8 & 63.6 & 66.5 & 54.3 & 62.1 & 59.5 & 24\\

    \hline
    
    \multirow{1}{*}{Voxel} &\multicolumn{1}{c|}{Cylinder3D~\cite{zhu2021cylindrical}} &  97.1 & 67.6 & 64.0 & 59.0 & 58.6 & 73.9 &  67.9 & 36.0 & 91.4 & 65.1 & 75.5 & 32.3 & 91.0 & 66.5 & 85.4 & 71.8 & 68.5 & 62.6 & 65.6 & 67.8 & 6 \\    

    \multirow{1}{*}{Projection} &\multicolumn{1}{c|}{PolarNet~\cite{zhang2020polarnet} *} &  93.8 & 40.3 & 30.1 & 22.9 & 28.5 & 43.2 & 40.2 & 5.6 & 90.8 & 61.7 & 74.4 & 21.7 & 90.0 & 61.3 & 84.0 & 65.5 & 67.8 & 51.8 & 57.5  & 54.3 & 16\\
    
    \multirow{1}{*}{Projection} &\multicolumn{1}{c|}{PolarNet+Ours} & \textbf{95.2} & \textbf{53.5} & \textbf{45.9} & \textbf{41.8} & \textbf{42.0} & \textbf{60.8} & \textbf{54.1} & \textbf{29.8} & \textbf{91.8} & \textbf{63.6} & \textbf{75.8} & \textbf{24.8} & \textbf{90.5} & \textbf{61.6} & 82.1 & 65.2 & 66.3 & \textbf{54.4} & \textbf{61.3} & 61.1 & 16\\    
    \hline
    \end{tabular}}
  \label{tab:exp}
\end{table*}

We use the original model structure and parameters of Cylinder3D\cite{zhu2021cylindrical} and PolarNet\cite{zhang2020polarnet}. Here, we list our main settings and parameters used in knowledge distillation. We select the 2nd and 3rd layers in the encoder and decoder for voxel-to-pillar distillation and the last layer for label-weight distillation. We divide the whole scene into 24 regions and set M to 2 for selective sampling for label-weight distillation.
For other parameters, we set the batchsize to 2 and use Adam\cite{Kingma2015adam} with a learning rate of 0.001 for optimization. We train our model for 40 epochs. The coefficients of loss balance are $2,2,1$.

\subsection{Results}
The results on the SemanticKITTI dataset~\cite{behley2019iccv} are shown in Tab.~\ref{tab:exp}. We compare different methods on the test set, including point-based methods such as KPConv~\cite{thomas2019kpconv} and RandLa-Net~\cite{hu2020randla} and projection-based methods such as RangeNet~\cite{milioto2019rangenet++}, SqueezeSegV2~\cite{wu2019squeezesegv2} and SalsaNext~\cite{cortinhal2020salsanext}. 
The last groups are our teacher model Cylinder3D~\cite{zhu2021cylindrical}, original PolarNet~\cite{zhang2020polarnet} and PolarNet trained with our framework. It can be seen that our model surpasses the baseline by 5\% mIoU. In Sec.~\ref{sec:method}, we mention that height information loss is the main reason for BEV-based methods performing worse than voxel-based methods. The experiment also proves that some classes, such as motorcycle, truck, and person, have a huge gap between Cylinder3D and PolarNet. Our proposed method mainly focuses on height information lost through voxel-to-pillar distillation and label-weight distillation. From the experiment, we can see that classes with richer height information perform better than the baseline.

We also conduct experiments on Like Paris-Lille-3D~\cite{roynard2017parislille3d}. The results shown in Tab.~\ref{tab:pl} demonstrate our method's effectiveness and generalization power.

\begin{table}[t]
\begin{center}
\caption{Results on Paris-Lille-3D~\cite{roynard2017parislille3d}.} 
\label{tab:pl}
\begin{tabular}{cccc}
  \hline
  {\makecell[c]{Model}} & {\makecell[c]{Categories}} & {\makecell[c]{mIoU}} & {\makecell[c]{FPS}}
  \\
  \hline
  Cylinder3D &  Voxel & 73.03 & 6\\
  PolarNet &  Projection & 64.04 & 13\\
  PolarNet+ours &  Projection &  65.71 & 13\\
  \hline
\end{tabular}
\end{center}
\end{table}

\subsection{Ablation Studies}

\begin{table}[t]
\begin{center}
\caption{Ablation studies for network components on the SemanticKITTI validation set. } 
\label{tab:components}
\begin{tabular}{c|cccc|c}
  \hline
  No. & PolarNet & Logit KD & VPD & LWD & mIoU
  \\
  \hline
  1 &  $\sqrt{ }$ &   &   &  & 57.08\\
  2 &  $\sqrt{ }$ & $\sqrt{ }$ &   &  & 57.31\\
  3 &  $\sqrt{ }$ & $\sqrt{ }$ & $\sqrt{ }$ &  &59.46\\
  4 &  $\sqrt{ }$ & $\sqrt{ }$ &   & $\sqrt{ }$ & 59.32\\
  5 &  $\sqrt{ }$ & $\sqrt{ }$ & $\sqrt{ }$ & $\sqrt{ }$ & 60.38\\
  \hline
\end{tabular}
\end{center}
\end{table}

\begin{table}[htbp]
\begin{center}
\caption{Ablation studies for different strategies for the voxel-to-pillar distillation module on the SemanticKITTI validation set. } 
\label{tab:vpd}
\begin{tabular}{c|ccc|c}
  \hline
  No.  & \makecell[c]{Compression\\ mode}  & \makecell[c]{Domain\\ Transfer} & \makecell[c]{Cross\\ Attention}& mIoU
  \\
  \hline
  1  & scatter-max &   &  & 58.05\\
  2  & sparse conv &   &  & 58.21\\
  3  & sparse conv & $\sqrt{ }$ &  &59.08\\
  4  & scatter-max & $\sqrt{ }$ & $\sqrt{ }$ & 59.32\\
  5  & sparse conv & $\sqrt{ }$ & $\sqrt{ }$ & 59.46\\

  \hline
\end{tabular}
\end{center}
\end{table}

In this part, we first perform ablation studies to show the effect of different modules in our framework. The results on the SemanticKITTI validation set are shown in Tab.~\ref{tab:components}. The results are improved after applying logit knowledge distillation between the classification logit of Cylinder3D~\cite{zhu2021cylindrical} and PolarNet~\cite{zhang2020polarnet}. Furthermore, voxel-to-pillar distillation and label-weight distillation, as the main components of our proposed framework, can improve the results by 2.38\% and 2.24\%, respectively. Finally, all these modules can improve the results on the validation set by approximately 3.3\% mIoU.

Then, we further explore the results of adopting different strategies in the voxel-to-pillar distillation module. We make some different variants of our model to show the effect of different settings. The result is shown in Tab.~\ref{tab:vpd}. We test different kinds of sparse operators, such as scatter-max and sparse convolution, used for changing the resolution of features, and we can see that sparse convolution is more appropriate to compress the 3D sparse features to the BEV plane. Domain transfer and cross-attention further improve the performance of our model. 



\subsection{Visualization}
\label{sec:visualization}

\begin{figure}[t]
\centering
\subfigure[Point clouds]{
		\label{sub1}
		\includegraphics[width=0.28\linewidth,height=0.25\linewidth]{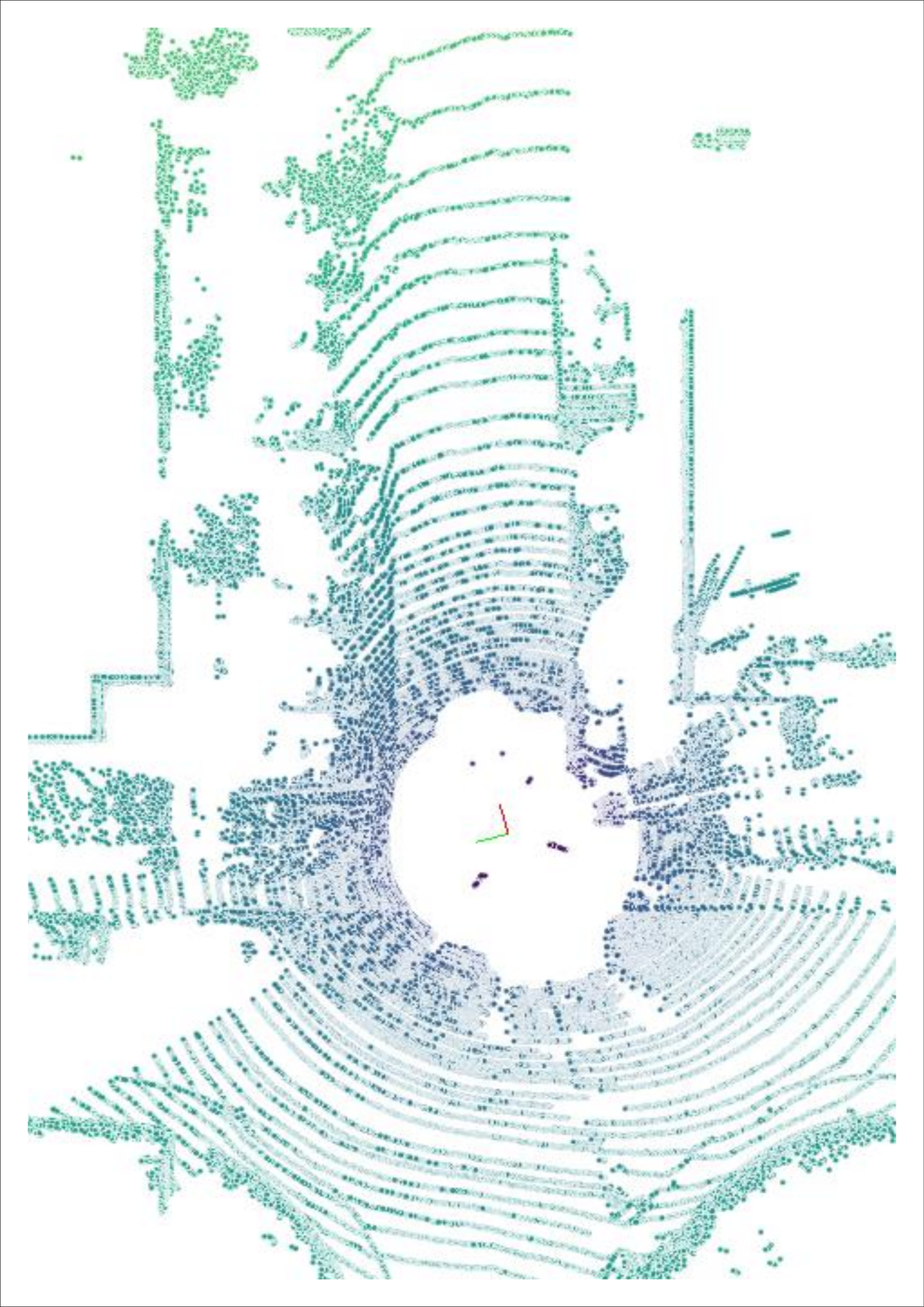}
}
\hfill
\subfigure[GT]{
		\label{sub2}
		\includegraphics[width=0.28\linewidth,height=0.25\linewidth]{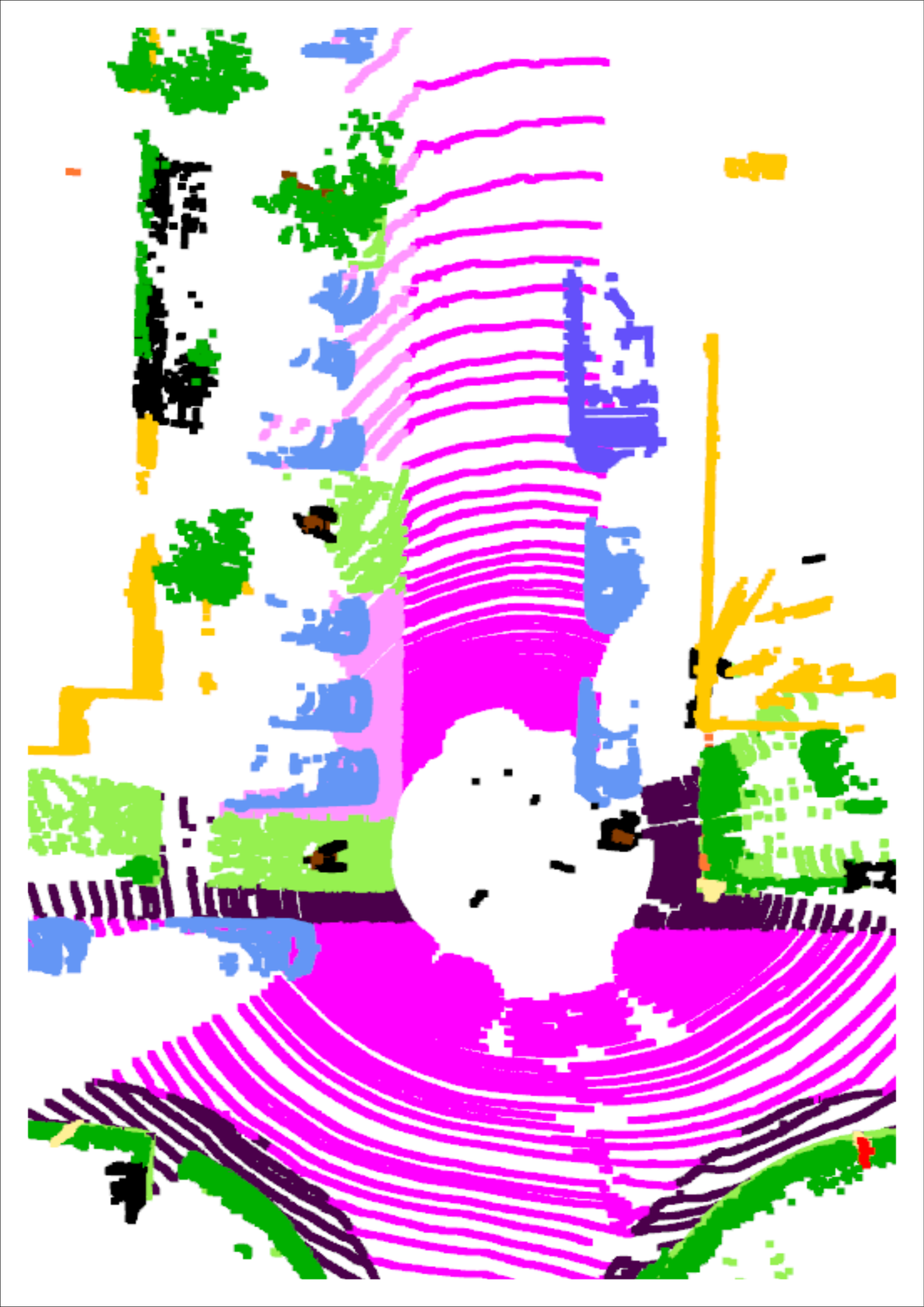}
}
\hfill
\subfigure[Height Map]{
		\label{sub3}
		\includegraphics[width=0.28\linewidth,height=0.25\linewidth]{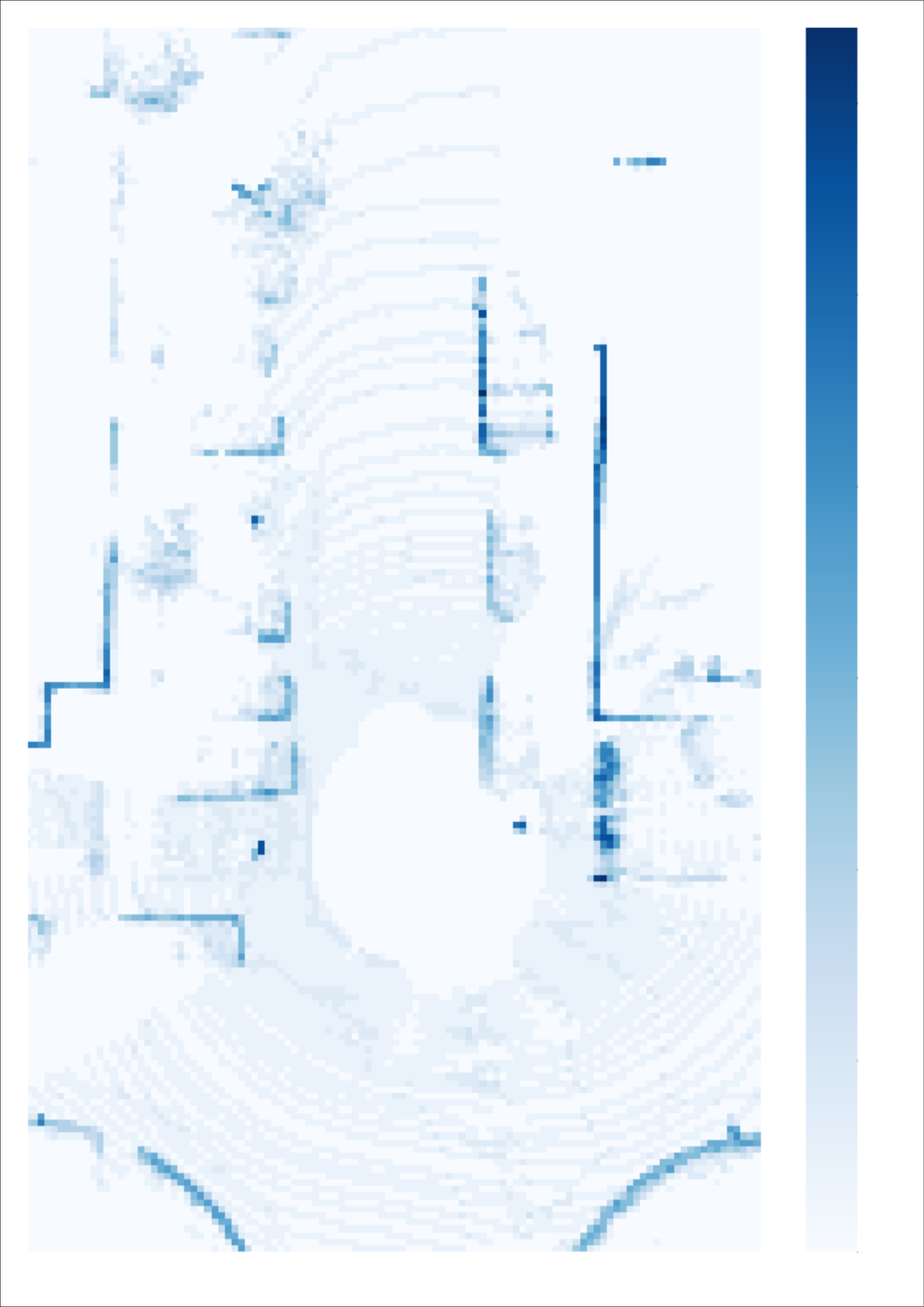}
}
\vfill
\subfigure[Cylinder3D]{
		\label{sub4}
		\includegraphics[width=0.28\linewidth,height=0.25\linewidth]{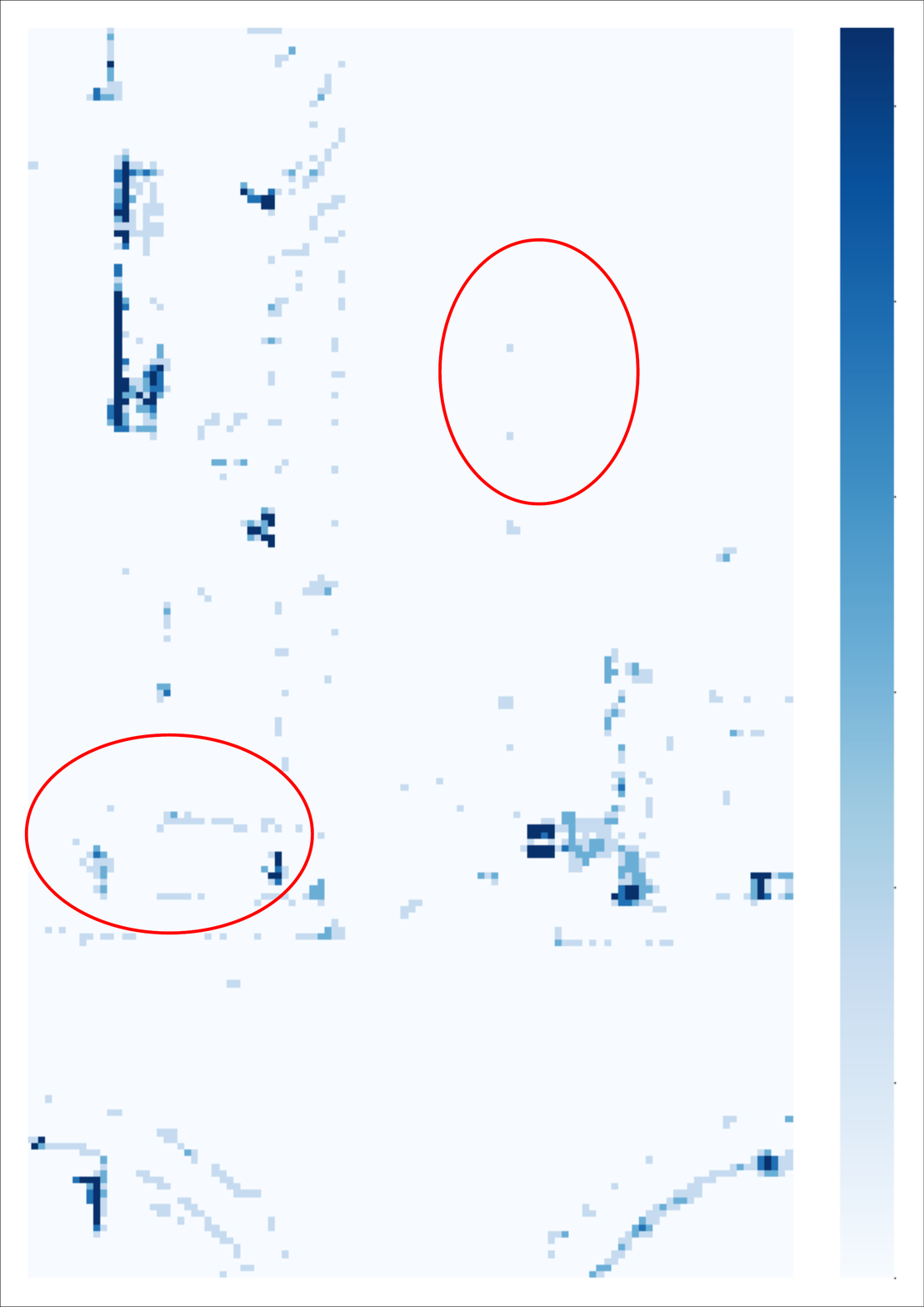}
}
\hfill
\subfigure[PolarNet]{
		\label{sub5}
		\includegraphics[width=0.28\linewidth,height=0.25\linewidth]{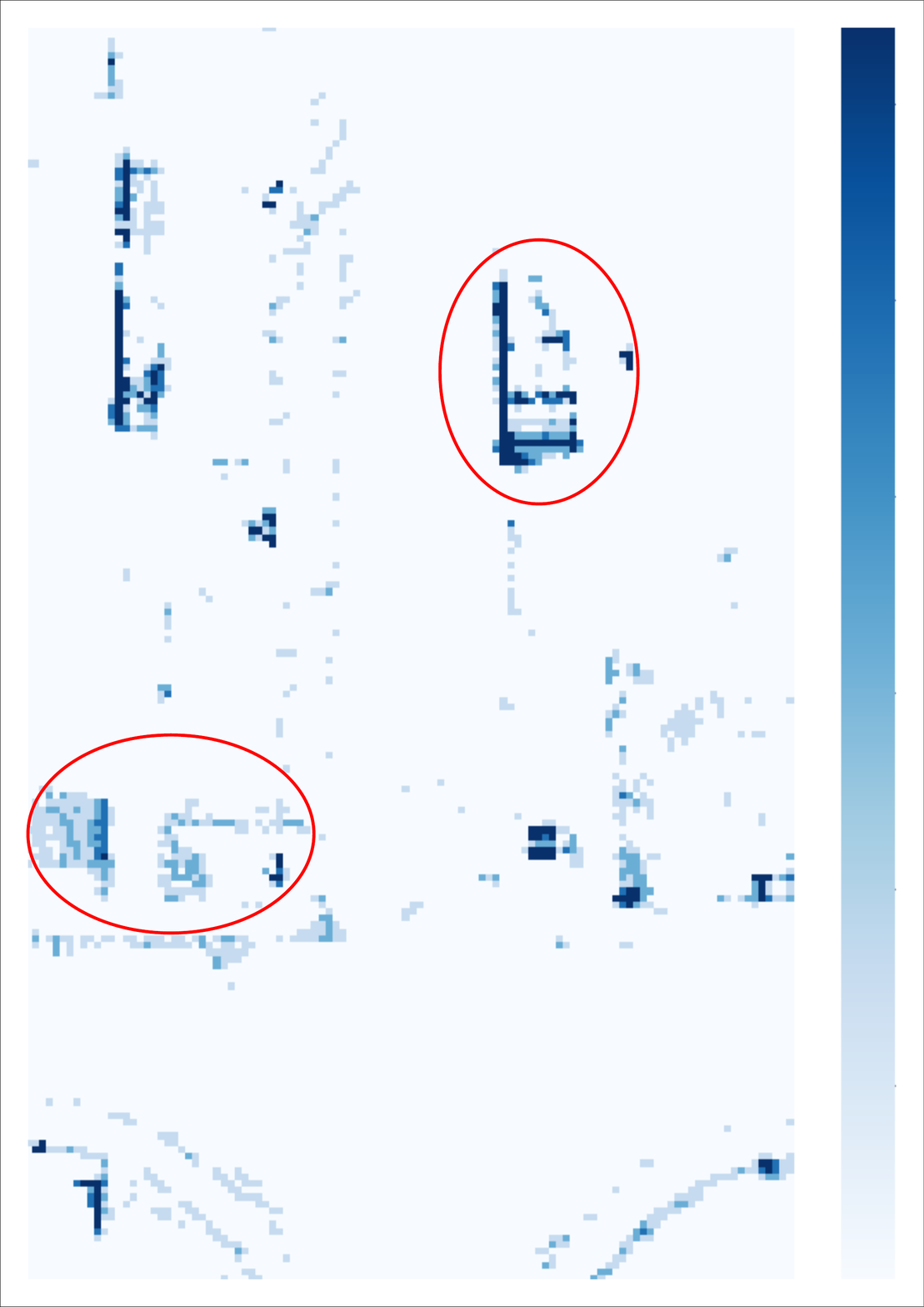}
}
\hfill
\subfigure[Ours]{
		\label{sub6}
		\includegraphics[width=0.28\linewidth,height=0.25\linewidth]{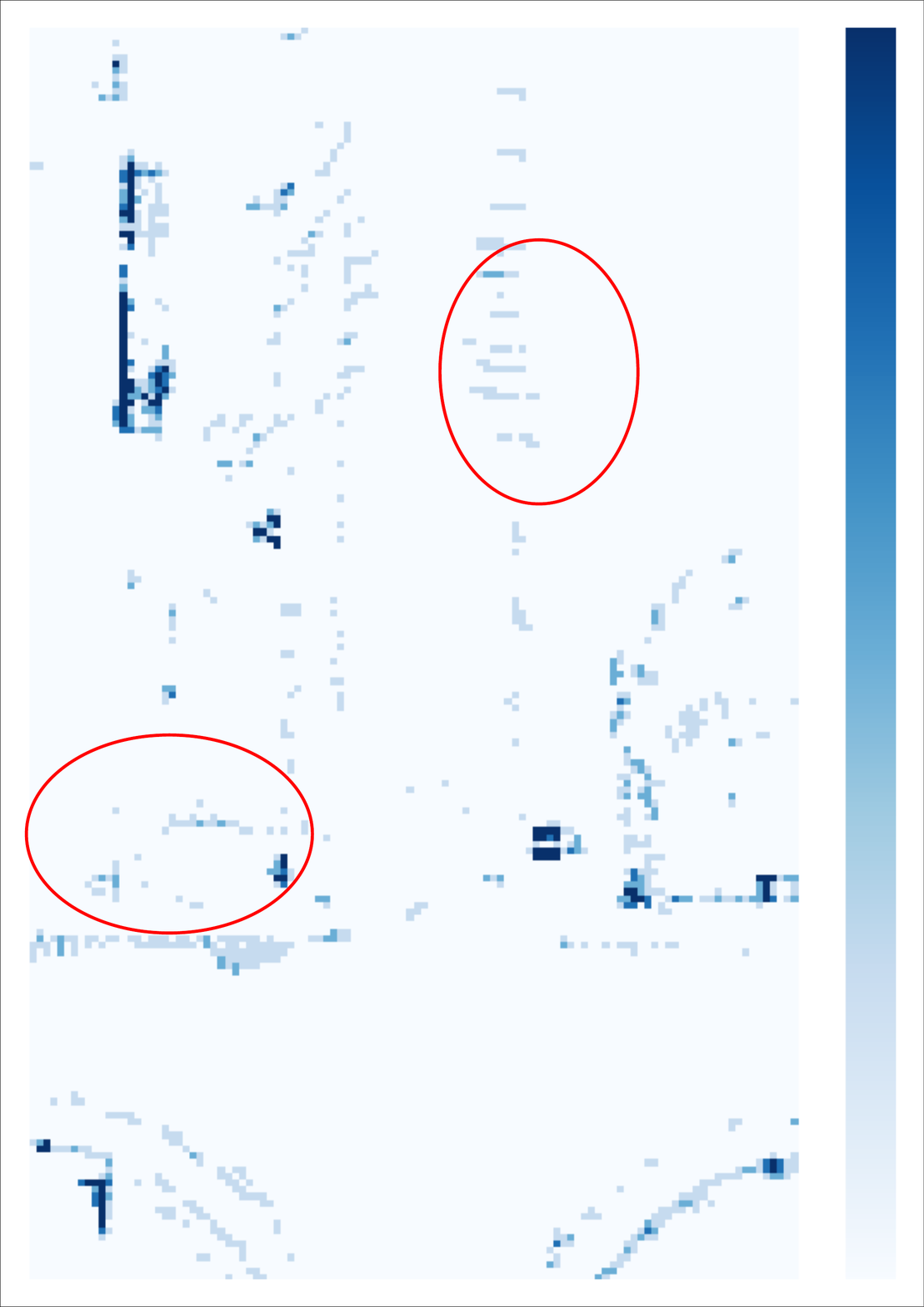}
}
\vfill
	\caption{Comparisons of false prediction maps of Cylinder3D, PolarNet and our method on the SemanticKITTI dataset~\cite{behley2019iccv}.}
	\label{visual}
\end{figure}

Fig.~\ref{visual} shows the visual comparison between Cylinder3D, PolarNet, and our method. We can see that the error numbers of different methods are different at locations with obvious height changes or rich semantic information. Fig.~\ref{sub1} and Fig.~\ref{sub2} are the point clouds and ground truth, and Fig.~\ref{sub3} is the number of non-empty voxels at different locations, where the darker the color is, the greater the height change. Fig.~\ref{sub4}, Fig.~\ref{sub5} and Fig.~\ref{sub6} are the error number counts at different locations of Cylinder3D~\cite{zhu2021cylindrical}, PolarNet~\cite{zhang2020polarnet} and our method. As shown in the red circles, our method can predict objects such as buildings, trucks, etc., as correctly as Cylinder3D, but PolarNet often makes incorrect predictions in this case.

\section{Conclusions}
In this paper, we propose a general knowledge distillation framework from voxel-based models to BEV-based models for point cloud semantic segmentation. 
Voxel-to-pillar distillation distills sparse 3D features to 2D BEV features for middle layers and makes the BEV-based model learn more structural and geometric knowledge. Label-weight distillation is used for the last layer before classification, which helps the model pay more attention to regions with more height information.
Experiments on the SemanticKITTI dataset and Paris-Lille-3D demonstrate that our method can outperform the baseline by 5\% mIoU on the test set, especially for classes such as motorcycle and person, with more than 15\% improvement.

\bibliographystyle{IEEEbib}
\small
\bibliography{icme2023template.bib}

\end{document}